\documentclass[runningheads]{llncs}
\usepackage{graphicx}
\usepackage{hyperref}
\usepackage{amsmath,amssymb}
\usepackage{mathtools}

\usepackage{tikz}
\usepackage{tikz-3dplot}
\usepackage{algorithm}
\usepackage{algpseudocode}

\algrenewcommand{\algorithmiccomment}[1]{// #1} 
\newcommand*\Let[2]{\State #1 $\gets$ #2} 

\DeclareMathOperator*{\argmin}{arg\,min}

\usepackage{xcolor}

\begin{document}

\pagestyle{headings}
\mainmatter

\title{Multigrid for Bundle Adjustment}
\author{Tristan Konolige \and
Jed Brown
}
\authorrunning{T. Konolige \and J. Brown}
\institute{University of Colorado Boulder}


\maketitle

\begin{abstract}
  Bundle adjustment is an important global optimization step in many structure from motion pipelines.
  Performance is dependent on the speed of the linear solver used to compute steps towards the optimum.
  For large problems, the current state of the art scales superlinearly with the number of cameras in the problem.
  We investigate the conditioning of global bundle adjustment problems as the number of images increases in different regimes and fundamental consequences in terms of superlinear scaling of the current state of the art methods.
  We present an unsmoothed aggregation multigrid preconditioner that accurately represents the global modes that underlie poor scaling of existing methods and demonstrate solves of up to 13 times faster than the state of the art on large, challenging problem sets.
\end{abstract}

\section{Introduction}

Bundle adjustment is a nonlinear optimization step often used in structure from motion (SfM) and SLAM applications to remove noise from observations.
Because such noise can have long range effects, it is necessary that this optimization step be global.
For large SfM problems, say reconstructing a whole city, the problem size can become very large.
As the problem size grows, existing techniques start to fail.
We introduce a new method that scales better than existing preconditioners on large problem sizes.

There are a variety of techniques to solve the bundle adjustment problem, but the most commonly used one is Levenberg-Marquardt~\cite{agarwal2010bundle,kushal2012visibility,triggs1999bundle}.
Levenberg-Marquardt is a iterative nonlinear least-squares optimizer that solves a series of linear systems to determine steps it takes towards the optimum.
Performance of Levenberg-Marquardt depends heavily on the performance of the linear system solver.
The linear system has a special structure created by the interaction of cameras and points in the scene.
The majority of entries in this system are zero, so sparse matrices are used.
Using the Schur complement, the linear system can be turned into a much smaller reduced system~\cite{triggs1999bundle}.
A common choice for solving the reduced system is either Cholesky for small systems, or iterative solvers for large systems~\cite{ceres-solver}.
The family of iterative solvers used, Krylov methods, have performance inversely related to the condition number of the linear system~\cite{toselli2005domain}.
This leads many to couple the Krylov method with a preconditioner: a linear operator that, when applied to the linear system, reduces the condition number.
There are a broad number of linear systems in the literature and hence a large number of preconditioners.
Preconditioners trade off between robustness and performance; preconditioners tailored to a specific problem are usually the fastest, but require an expert to create and tune.

Preconditioners already used in the literature are point block Jacobi~\cite{agarwal2010bundle}, successive over relaxation~\cite{agarwal2010bundle}, visibility-based block Jacobi~\cite{kushal2012visibility}, and visibility-based tridiagonal~\cite{kushal2012visibility}.
Of these, the visibility based preconditioners are the fastest on large problems.
We propose a new multigrid preconditioner that outperforms point block Jacobi and visibility based preconditioners on large, difficult problems.
Multigrid methods are linear preconditioners that exploit multilevel structure to scale linearly with problem size.
Multigrid methods originated from the need to solve large systems of partial differential equations, and there has been some success applying multigrid to non-PDE areas like graph Laplacians~\cite{lamg}.
Our multigrid methods exploits the geometric structure present in bundle adjustment to create a fast preconditioner.

\section{Background}

\subsection{Bundle Adjustment}
%

Bundle adjustment is a nonlinear optimization problem over a vector of camera and point parameters $x$ with goal of reducing noise from inaccurate triangulations in SfM.
We use a nonlinear least-squares formulation where we minimize the squared sum of reprojection errors~\cite{triggs1999bundle}, $f_i$, over all camera-point observations,
$$ x^* = \argmin_{x} \sum_{i\in\text{observations}} \left\lVert f_i(x) \right\rVert.$$
Here $\sum || f_i ||$ is the \textit{objective function}.
A usual choice of solver for the nonlinear least-squares problem is the Levenberg-Marquardt algorithm \cite{levenberg1944method}.
This is a quasi-Newton method that repeatedly solves
$$\left(J_x^T J_x + D\right) \delta_i = - J_x f(x),\quad J_x = \frac{\partial f}{\partial x}, $$
where $D$ is a diagonal damping matrix, to compute steps $\delta_i$ towards a minimum.
Levenberg-Marquardt can be considered as a combination of Gauss-Newton and gradient descent.


Solving the linear system, $J^T J + D$, is the slowest part of bundle adjustment.
Splitting $x$ into $[x_c x_p]^T$, where $x_c$ are the camera parameters and $x_p$ are the point parameters, yields a block system
\begin{align}
  F &= J_{x_c},\quad E = J_{x_p},\quad J_x = \begin{bmatrix} F & E \end{bmatrix},\\
  J_x^TJ_x+D &= \begin{bmatrix}
    A = F^T F + D_{x_c} & F^T E       \\
    E^T F       & C = E^T E + D_{x_p} \\
  \end{bmatrix}.
\end{align}
$C$ is a block diagonal matrix with blocks of size $3 \times 3$ corresponding to point parameters.
$A$ is a block diagonal matrix with blocks of size $9 \times 9$ corresponding to camera parameters.
$E^TF$ is a block matrix with blocks of size $9 \times 3$ corresponding to interaction between points and cameras.
A usual trick to apply is using the Schur complement to eliminate the point parameter block $C$:
$$
S = A - F^TE C^{-1} E^T F.
$$
$S$ is block matrix with blocks of size $9 \times 9$.
$C$ is chosen over $A$ because the number of points is often orders of magnitude larger than the number of cameras.
Thus, applying the Schur complement greatly reduces the size of the linear system being solved.

The Schur complement system has structure determined by the covisibility of cameras: if $c_i$ and $c_j$ both observe the same point, then the block $S_{ij}$ is nonzero.
In almost all scenarios, cameras do not have points in common with every other camera, so $S$ is sparse.
$S$ tends to be easier for the linear solver when all cameras view the same object, for example, tourist photos of the Eiffel Tower.
Cameras are close together and a single camera out of place has little effect on the other cameras (because of the high amount of overlap between views).
On the other hand, linear solvers are slower when cameras view a large area, like in \textit{street view} where images taken from a car as it drives around a city.
In this situation, adjusting a single camera's position has an effect on all cameras near to it, causing long dependency chains between cameras.
Long dependency chains cause issues for iterative linear solvers as information can only be propagated one step in the chain per iteration of the solver.

This linear system is normally not solved to a tight tolerance.
Usually, a fairly inexact solve of the linear problem can still lead to good convergence in the nonlinear problem \cite{agarwal2010bundle}.
As the nonlinear problem gets closer to a minima, the accuracy of the linear solve should increase.
The method for controlling the linear solve accuracy is called a \textit{forcing sequence}.
Ceres Solver \cite{ceres-solver}, the current state of the art nonlinear least-squares solver, uses a criteria proposed by Nash and Sofer \cite{nash1990assessing} to determine when to stop solving the linear problem:
\begin{align}
  Q_n = \frac{1}{2} x^T A x - x^T b, \\
  \text{stop if  } i \frac{Q_i - Q_{n-1}}{Q_i} \leq \tau,
\end{align}
where $i$ is the current conjugate gradients iteration number and $\tau$ is the tolerance from the forcing sequence.
It is important to note that the occurrence of the iteration number in the criteria means that more powerful preconditioners end up solving the linear problem to a tighter tolerance.

When using a simple projective model, the bundle adjustment problem is ill-conditioned.
Causes for ill-conditioning include difference in scale between parameters and a highly nonlinear distortion.
Improving the conditioning of the problem is possible, for example, in~\cite{qu2018efficient}, Qu adaptively reweights the residual functions and uses a local parameterization of the camera pose to improve conditioning.
Changes like this are orthogonal to improving linear solver performance, so we use a simple projective model for this paper.

\subsection{Existing Solvers}

There are a variety of ways to solve $S$.
Konolige uses a sparse direct Cholesky solver~\cite{konolige2010sparse}.
Sparse direct solvers are often a good choice for small problems because of their small constant factors.
For large problems with 2D/planar connectivity, sparse direct methods require $O(n^{1.5})$ time and $O(n\log n)$ space when small vertex separators exist (a set of vertices whose removal splits the graph in half) \cite{lipton1979generalized}.
In street view problems, camera view overlap in street intersections creates large vertex separators, making sparse direct solvers a poor choice for large problems.
To improve scaling, Agarwal et al. propose using conjugate gradients with Jacobi preconditioning~\cite{agarwal2010bundle}.
Kushal and Agarwal later extend this work with block-Jacobi and block-tridiagonal preconditioners formed using the \textit{visibility}, or number of observed points in common between cameras~\cite{kushal2012visibility}.
Jian et al. propose using a preconditioner based off of a subgraph of the unreduced problem similar to a low-stretch spanning tree \cite{jian2012generalized}.

\subsection{Algebraic Multigrid}

\begin{figure}
  \centering
  \resizebox{\linewidth}{!}{\includegraphics{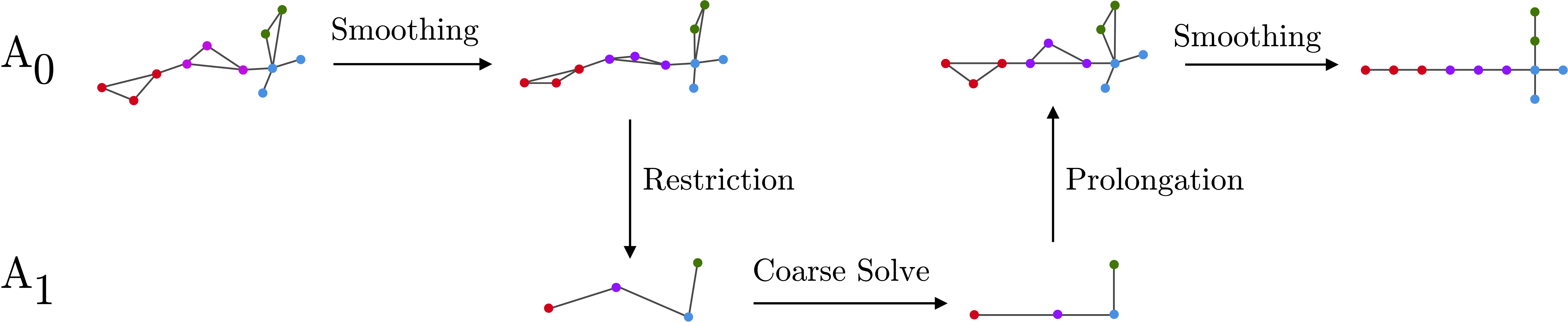}}
  \caption{
    Steps in multigrid to solve a linear system.
    Points represent degrees of freedom and lines represent interactions between them.
    Smoothing reduces local error, while the coarse solve along with restriction and prolongation reduce long range errors.
  }
  \label{fig:multigrid}
\end{figure}


Algebraic Multigrid (AMG) is a technique for constructing scalable preconditioners for symmetric positive definite matrices.
AMG constructs a series of increasingly smaller matrices $\{A_0: \mathbb{R}^{n_0\times n_0}, A_1: \mathbb{R}^{n_1\times n_1}, ...\}$ that are approximations to the original matrix $A_0$.
$A_0$ is solved by repeatedly solving coarse levels, $A_{l+1}$, and using the solution on the fine level, $A_l$.
The restriction ($R_l:\mathbb{R}^{n_{l+1}\times n_l}$) and prolongation ($P_l:\mathbb{R}^{n_{l}\times n_{l+1}}$) matrices map from $A_{l}$ to $A_{l+1}$ and back, respectively.
The coarse solve accurately corrects error in long range interactions on the fine level.
This coarse level correction is paired with a \textit{smoother} that provides local correction.
The combination of coarse grid correction and fine grid smoothing, when applied to the entire \textit{hierarchy} of levels ($\{A_0, A_1, ...\}$), creates a preconditioner that bounds the iteration count of the iterative solver independently of problem size.
Algorithm~\ref{alg:vcycle} shows a full multigrid preconditioner application (one \textsc{mgcycle} is one preconditioner application).

\begin{algorithm}
  \caption{Multigrid Preconditioner Application}
    \label{alg:vcycle}
  \begin{algorithmic}[1]
    \Function{mgcycle}{level $l$, initial guess $x$, rhs $b$}
      \If{$l$ is the coarsest level}
        \Let{$x$}{Direct solve on $A_{l} x = b$}
        \State \Return{$x$}
      \Else
        \Let{$x$}{smooth($x, b$)} \Comment{Pre-smoothing} \label{line:presmooth}
        \Let{$r$}{$b - A_l x$} \Comment{Residual} \label{line:residual}
        \Let{$r_c$}{$R_l r$} \Comment{Restriction} \label{line:restriction}
        \Let{$x_c$}{$0$}
        \Let{$x_c$}{\Call{mgcycle}{$l+1, x_c, r_c$}} \Comment{Coarse level solve} \label{line:coarse}
        \Let{$x$}{$x + P_l x_c$} \Comment{Prolongation} \label{line:prolongation}
        \Let{$x$}{smooth($x, b$)} \Comment{Post-smoothing} \label{line:postsmooth}
        \State \Return{$x$}
      \EndIf
    \EndFunction
  \end{algorithmic}
\end{algorithm}

Multigrid performance depends on the choice of smoother and method of constructing the coarse grid.
Usual choices of smoother are point block Jacobi, point block Gauss-Seidel, and Chebyshev.
Typically one or two iterations of the smoother are applied for pre- and post-smoothing.
Smoothers must reduce local error and be stable on long range error.
Aggregation based methods construct $R$ and $P$ by partition the degrees of freedom of the fine level into non-overlapping \textit{aggregates}~\cite{vanek1996algebraic}.
Each aggregate corresponds to a single degree of freedom on the coarse level.
$R$ computes each aggregate's coarse level dof as a weighted average of all the aggregate's dofs on the fine level.
$P$ applies the same process in reverse, so $P = R^T$.
Given a level $A_{l}$, the coarse level matrix is constructed as $A_{l+1}=R_l A_l P_l$.
Choosing aggregates is problem dependent, and is an important contribution of our paper.

\section{The Algorithm}

\subsection{Nullspace}\label{sec:nullspace}

Fast convergence of multigrid requires satisfaction of the \textit{strong approximation property},
$$\min_u||e-Pu||^2_A \leq \frac{\omega}{||A||} \langle Ae,Ae \rangle,$$
for some fine grid error $e$ and constant $\omega$ determining convergence rate \cite{tamstorf2015smoothed,rugestuben,maclachlan2014theorectical}.
To satisfy this condition, $e$s for which $||Ae||$ is small (\textit{near-nullspace} vectors) must be accurately captured by $P$.
For any bundle adjustment formulation with monocular cameras and no fixed camera locations, $J^TJ$ has a nullspace, $N$, with dimension 7 corresponding to the free modes of the nonlinear problem~\cite{triggs1999bundle}.
These are 3 rotational modes, 3 translational modes, and 1 scaling mode.
When the damping matrix $D$ is small, this nullspace becomes near-nullspace vectors, $K$, of $J^TJ+D$.
For the Schur complement system, the near-nullspace of $S$ is $K_{x_c}$.
We augment $K$ with 9 columns that are constant on each of the 9 camera parameters.


\subsection{Aggregation}

\begin{algorithm}
  \caption{Aggregation}
    \label{alg:aggregation}
  \begin{algorithmic}[1]
    \Function{aggregate}{strength of connection matrix $G$ of size $n \times n$}
      \For{$i \in 1..n$}
        \If{$i$ is unaggregated}
          \For{$j \in G_{i,:}$ sorted by largest to smallest}
            \If{$j$ is unaggregated}
              \State form new aggregate with $i$ and $j$
              \State break
            \ElsIf{$j$ is in aggregate $k$ and $\text{size}(k) < 20$}
              \State add $i$ to aggregate $k$
              \State break
            \EndIf
          \EndFor
        \EndIf
      \EndFor
    \EndFunction
  \end{algorithmic}
\end{algorithm}

The multigrid aggregation algorithm determines both how quickly the linear system solver converges and the time it takes to apply the preconditioner.
Choosing aggregates that are too large results in a cheap cycle that converges slowly.
On the other hand, if aggregates are too small, the solver will converge quickly but each iteration will be computationally slow.
The aggregation routine needs to strike the right balance between too large and too small aggregates.

Typical aggregation routines for multigrid form fixed diameter aggregates by clustering together a given ``root'' node with all its neighbors.
This technique works well on PDE problems where the connectivity is predictable and each degree of freedom has is connect to a limited number of other degrees of freedom.
Bundle Adjustment does not necessarily have these characteristics.
Street view-like problems might have low degree for road sections that do not overlap, but when roads intersect, some dof's can be connected to many others.
Choosing one of these well connected dof's as the root of an aggregate creates a too aggressive coarsening.

Aggregation routines for non-mesh problems exist, for example, for graph Laplacians \cite{lamg,ligmg}.
These routines have to contend with dofs that are connected to a majority of other dofs; something we do not expect to see in street view-like datasets.
Instead, we use a greedy algorithm that attempts to form aggregates by aggregating unaggregated vertices with their ``closest'' connected neighbor and constrains the maximum size of aggregates to prevent too aggressive coarsening.

Closeness of dofs is determined by the \textit{strength of connection} matrix.
Almost all multigrid aggregation algorithms use this matrix as an input to determine which vertices should be aggregated together.
The strength of connection matrix can be created based only using matrix entries (for example, the affinity \cite{lamg} and algebraic distance metrics \cite{brandt2015algebraic}) or use some other, geometric information.
For bundle adjustment, this other information can be camera and point positions or the visibility information between them.
The strength of connection metric we choose to use is the visibility metric used by Kushal and Agarwal in \cite{kushal2012visibility}.
We tried other metrics, like including the percentage of image overlap between two cameras, but the visibility metric remained superior.
The visibility strength of connection matrix, $G$, is defined as:
\begin{align}
  G_{i,j} &= \begin{cases}
    0 & i == j, \\
    \frac{v_i^T v_j}{||v_i|| ||v_j||} & \text{otherwise},
  \end{cases} \\
  v_{k_l} &= \begin{cases}
    1 & \text{ camera } k \text{ sees point } l, \\
    0 & \text{ otherwise}.
  \end{cases}
\end{align}

Most aggregation routines for PDE's enforce some kind of diameter constraint on aggregates.
We find that for our problems this is not necessary.
However, we force aggregates to contain no more than 20 dofs, to ensure our aggregates do not become very large.
In practice, we see that aggregate sizes are usually in the range for 8 to 3, with the mean aggregate size usually just a little more than 3.

\subsection{Prolongation}

We use a standard multigrid prolongation construction technique~\cite{adams2002evaluation,tamstorf2015smoothed}.
For each aggregate, the nullspace is restricted to the aggregate, a QR decomposition is applied, and $\mathcal{Q}$ becomes a block of $P$ while $\mathcal{R}$ becomes a block of the coarse nullspace:
\begin{align}
  \mathcal{Q}_{\text{agg}}\mathcal{R}_{\text{agg}} &= K_{\text{agg}} \text{ forall agg} \in \text{ aggregates}, \\
  P &= \Pi\begin{bmatrix}
    \mathcal{Q}_1 &   &   \\
      & \ddots &   \\
      &   & \mathcal{Q}_n
  \end{bmatrix}, \\
  K_{\text{coarse}} &= \Pi\begin{bmatrix}
    \mathcal{R}_1 \\
    \vdots \\
    \mathcal{R}_n
  \end{bmatrix}.
\end{align}
Here $\Pi$ is a permutation matrix from contiguous aggregates to the original ordering.
Using the QR decomposition frees us from having to compute the local nullspace and represent it on the coarse level (this would require computing the center of mass of each aggregate).
Our near-nullspace has dimension 16 (7 from $J^TJ$'s nullspace, 9 from per dof constant vectors), so each of our coarse level matrices has 16 by 16 blocks.

\subsection{Smoother}

We use a Chebyshev smoother~\cite{adams2003parallel} with a point block Jacobi matrix.
We find the Chebyshev smoother to be more effective than block-Jacobi and Gauss-Seidel smoothing.
The Chebyshev smoother does come with a disadvantage: it requires an estimate of the largest eigenvalue, $\lambda_{\text{max}}$, of $D^{-1}A$.
Like Tamstorf et al., we find that applying generalized Lanczos on $Ax=\lambda D x$ is the most effective way to find the largest eigenvalue \cite{tamstorf2015smoothed}.
This eigenvalue estimate is expensive, so we limit it to 5 applications of the operator.
We use $1.1\lambda_{\text{max}}$ for the high end of the Chebyshev bound and $0.3\lambda_{\text{max}}$ as the lower end.
However, the superior performance of the Chebyshev smoother outweighs its increased setup cost.
We also tried using the Gershgorin estimate of the largest eigenvalue, but that proved to be very inaccurate (by multiple orders of magnitude).
We apply two iterations of pre-smoothing and two of post-smoothing.

\subsection{To Smooth or Not To Smooth}

Aggregation-based multigrid uses prolongation smoothing in order to improve convergence \cite{vanek1996algebraic}.
Smoothing the prolongation operator is sufficient to satisfy the strong approximation property and achieve constant iteration count regardless of problem size.
Usually, smoothing the prolongation operator improves convergence rate at the cost of increased complexity of the coarse grids.
In PDEs and other problems with very regular connectivity, this trade off is worthwhile.
However, in other problems, like in irregular graph Laplacians, irregular problem structure causes massive fill-in---coarse grids become dense~\cite{lamg}.
The street view bundle adjustment problems we are working with appear to be similar in structure to PDE based problems: the number of nonzeros per row is bounded and the diameter of the problem is relatively large.
However, when we apply prolongation smoothing to our multigrid preconditioner, we see large fill-in in the coarse grids, similar to what happens in irregular graph Laplacians.
Although the nonzero structure of street view bundle adjustment appears similar to PDEs, it still has places where dofs are coupled with many other dofs---places where large fill-in occurs.
These places could be landmarks that are visible from far away or intersections where there is a large amount of camera overlap.
The large fill-in on the coarse grid makes the setup phase too expensive to justify the improved performance in the solve phase.
Choosing not to smooth aggregates means our preconditioner does not scale linearly, but it does scale better than any of the current state of the art preconditioners.

\subsection{Implicit Operator}

On many bundle adjustment problems, it is often faster to apply the Schur complement in an implicit manner, rather than constructing $S$ explicitly~\cite{agarwal2010bundle}.
That is, we can apply manifest $Sx$ for a vector $x$ as $Ax - F^T(E(C^{-1}(E^T(Fx))))$.
As conjugate gradients requires only matrix-vector products, we can use the implicit matrix product with it for improved performance.
An issue arises when we use a preconditioner with CG: the preconditioner often needs the explicit representation of $S$.
For block-Jacobi preconditioning, Agarwal et al.~\cite{agarwal2010bundle} construct on the relevant blocks of $S$.
The same technique is used by Kushal and Agarwal in their visibility-based preconditioner~\cite{kushal2012visibility}.
For Algebraic multigrid, the explicit matrix representation is needed to form the Galerkin projection $P^T S P$.
We could use the implicit representation with the Galerkin projection, $P^T (A (P x)) - P^T (F^T (E (C^{-1}(E^T(F(Px))))))$, but then we are paying the cost of the full implicit matrix at each level in our hierarchy.
We instead compute the cost of using the implicit vs explicit product on each level of our hierarchy and choose the cheapest one.
In our tests we create both the implicit and explicit representations for each level, but only use the most efficient one (computing the number of nonzeros in a sparse matrix product is difficult without forming the product itself).
It would be possible to create only the needed representation on each level, but we have not explored the costs.
This may actually have a large performance impact as the Galerkin projection requires an expensive matrix-triple product (currently the most expensive part of setup).


\section{Generating Synthetic Datasets}

\begin{figure}
  \centering
  \resizebox{!}{4cm}{\includegraphics{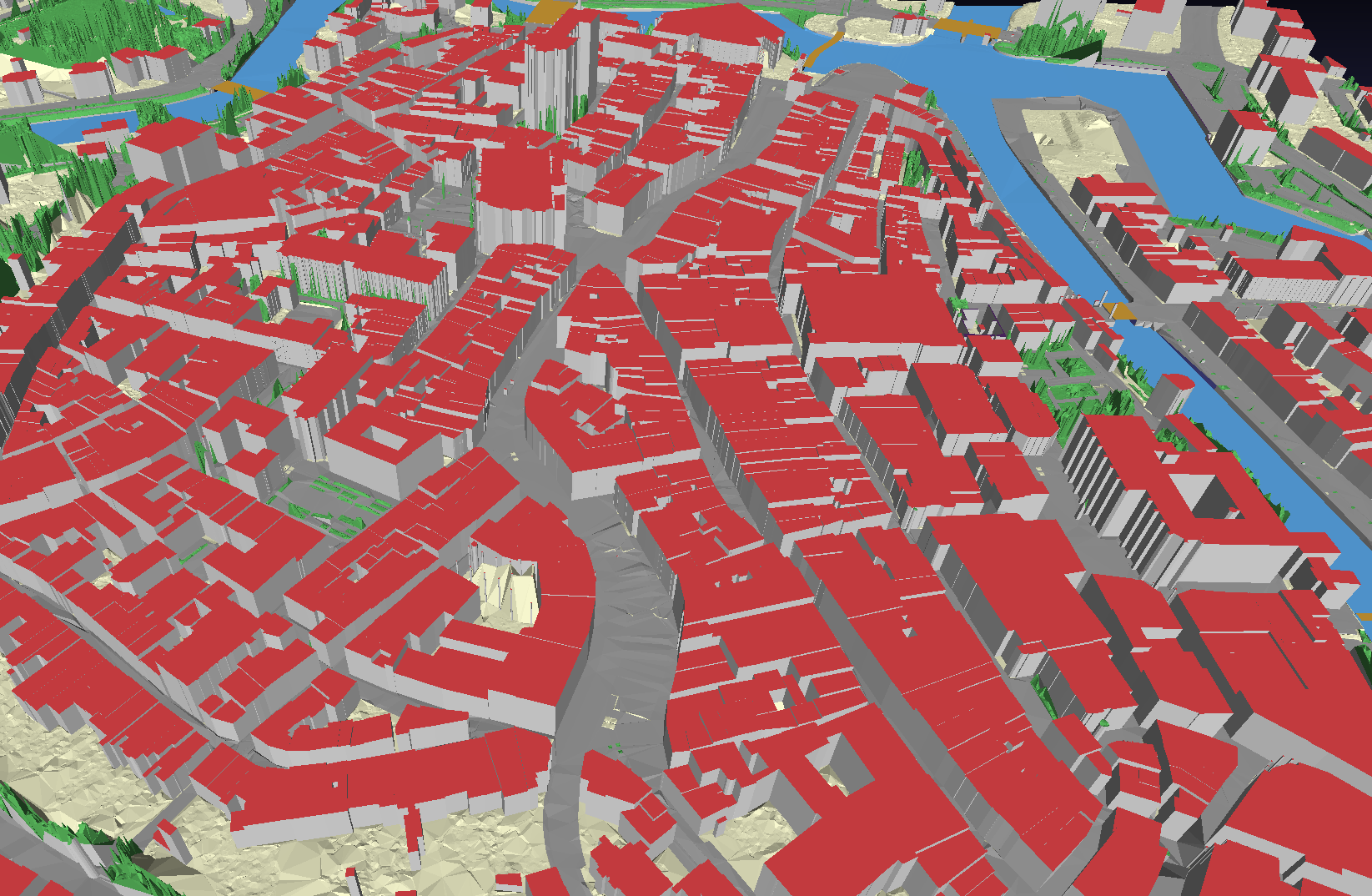}\includegraphics{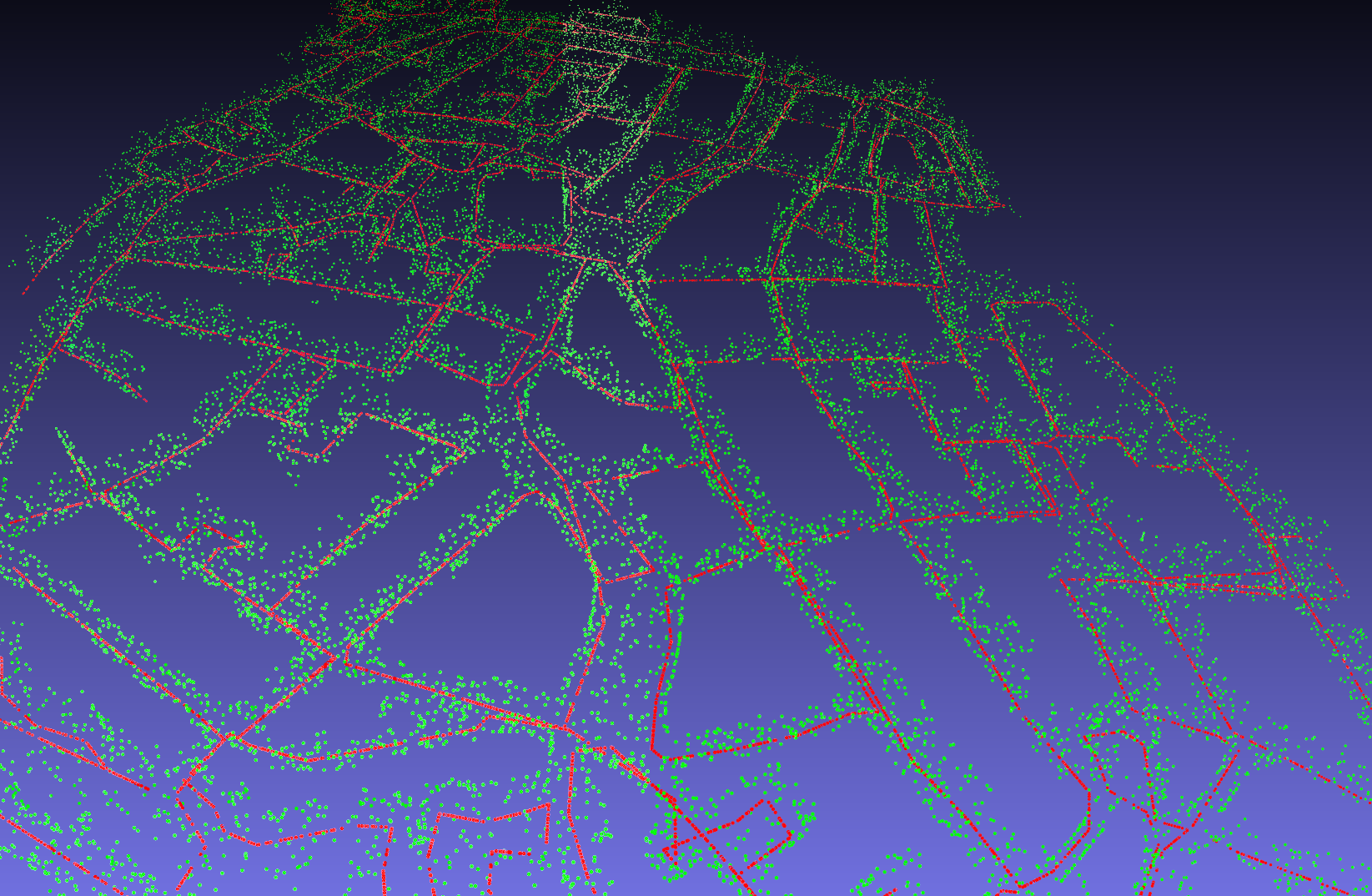}}
  \caption{
    Left: 3D model of Zwolle (Netherlands).
    Right: generated synthetic bundle adjustment dataset.
  }
  \label{fig:synthetic}
\end{figure}

To compare our preconditioner to existing ones, we need datasets to test against.
Only a couple of real-world datasets are publicly available, namely the Bundle Adjustment in the Large datasets\footnote{\href{http://grail.cs.washington.edu/projects/bal}{http://grail.cs.washington.edu/projects/bal}}~\cite{agarwal2010bundle} and the 1DSFM datasets\footnote{\href{http://www.cs.cornell.edu/projects/1dsfm}{http://www.cs.cornell.edu/projects/1dsfm}}~\cite{wilson2014robust}.
The largest of these datasets contains 15 thousand cameras.
As we are interested in evaluating scaling of our algorithms, we require much larger datasets.
Also most of these datasets are ``community photo'' style, i.e. there are many pictures of the same object.
Furthermore, all of these datasets contain too many outliers: long range effects are not exposed to the linear solver.
We would like datasets with more varied camera counts and visibility structure similar to what we would expect from street view, so we generate a series of synthetic datasets with these properties.


We generate a ground truth (zero error) bundle adjustment dataset by taking an existing 3D model of a city and drawing potential camera paths through it.
We generate random camera positions on these paths, then generate random points on the geometry and test visibility from every camera to every point.
We can control the number of cameras and the number of points to generate datasets of different sizes.
By choosing different 3D models or different camera paths, we can change the visibility graph between cameras and points.
For our datasets, we use a 3D model of Zwolle in the Netherlands\footnote{\href{https://3d.bk.tudelft.nl/opendata/3dfier/}{https://3d.bk.tudelft.nl/opendata/3dfier/}}.
Figure~\ref{fig:synthetic} shows the 3D model and one of the more complicated datasets we generated.

These datasets do not contain any error, so we add noise into each.
The straight forward approach of adding Gaussian noise directly to the camera and point parameters results in a synthetic problem that is easier to solve than the real world problems as it contains no non-local effects.
Instead, we add long rang drift to the problem: as cameras and points get farther from the origin we perturb their location more and more in a consistent direction.
Adding a little noise to the camera rotational parameters also helps as rotational error is highly nonlinear.
We are careful to not add too much noise or too many incorrect correspondences as this leads to problems with many outliers (see section~\ref{sec:robust} a solution).

\section{Results}

We tested our multigrid preconditioner against point block Jacobi and visibility-based block Jacobi preconditioners on a number of synthetic problems (we found the visibility-based tridiagonal preconditioner to perform similarly to visibility-based block Jacobi, so we omit it).
Our test machine is an Intel Core i5-3570K running at 3.40GHz with 16GB of dual-channel 1600MHz DDR3 memory.
For large problems, we use NERSC's Cori---an Intel Xeon E5-2698 v3 2.3 GHz Haswell processor with 128 GB DDR4 2133 MHz memory.
We use Ceres Solver~\cite{ceres-solver} to perform our nonlinear optimization as well as for the conjugate gradient linear solver.
Ceres Solver also provides the point block Jacobi and visibility based preconditioners.
We terminate the nonlinear optimization at 100 iterations or if any of Ceres Solver's default termination criteria are hit.
Our initial trust region radius is 1e4.
We use a constant forcing sequence with tolerance $\tau$.
Results are post processed to ensure that all nonlinear solves for a given problem end at the same objective function value.
For some preconditioners (like our multigrid), this significantly impacted the total number of nonlinear iterations taken (see section~\ref{sec:solver-accuracy}).

\begin{figure}
  \centering
  \resizebox{!}{6cm}{\input{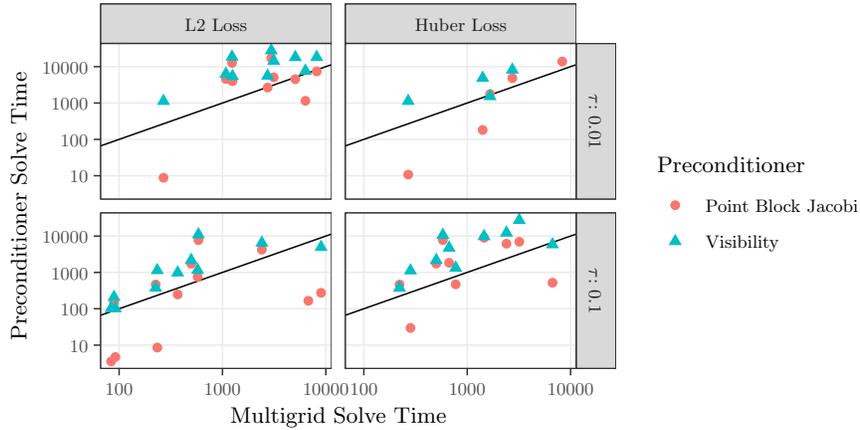}}
  \caption{
    Preconditioner solve time versus multigrid solve time for a set of synthetic problems with varying number of cameras, visibility structure, and noise.
    Solve time is measured as total time spent in the linear solver (setup and solve) for all nonlinear iterations to a certain problem dependent tolerance.
    Points above the diagonal (black line) indicate the problem was solved quicker with multigrid than the given preconditioner, points below indicate that multigrid was slower.
    Vertical columns of plots have use the same loss function.
    Horizontal plot rows have the same linear solve tolerance $\tau$.
    For the majority of cases, multigrid performs better than all the other preconditioners.
  }
  \label{fig:reltime}
\end{figure}

Our preconditioner is written in Julia~\cite{julia} and uses SuiteSparse~\cite{suitesparse} for its Galerkin products.
We have not spent much time optimizing our preconditioner.
We do not cache the sparse matrix structure between nonlinear iterations and reallocate almost all matrix products.
Furthermore, the Julia code allocates more and is slower than it could be if written in C or C++.
Jacobian matrices are copied between Julia and Ceres Solver, leading to a larger memory overhead.
We do not use a sparse matrix format that exploits the block structure of our matrices or use matrix-multiples that exploit this structure.
All of this is to say that our method could be optimized further for potentially greater speedup.

Still, our multigrid preconditioner performs better than point block Jacobi and visibility based preconditioners on most large problems.
Figure~\ref{fig:reltime} shows the relative solve time of other preconditioners vs our multigrid preconditioner for a variety of synthetic problems.
Our preconditioner is up to 13 times faster than point block Jacobi, and 18 times faster than visibility based preconditioners.
Median speedup is 1.7 times faster than point block Jacobi, and 2.8 times faster than visibility based preconditioners.
This includes cases where problems are large, but not difficult; a situation where our preconditioner performs poorly.
On smaller problems (with fewer than 1000 cameras), our preconditioner is significantly slower than direct methods.

On problems where the geometry is simpler, point block Jacobi normally outperforms visibility based preconditioners and our preconditioner.
This is because the linear problems are relatively easier to solve and the visibility based methods cannot recoup their expensive setup cost.
On more complicated problems (when the camera path crosses itself), the difficulty of the solve makes the high setup cost of the visibility based methods worthwhile.
We find that these more complicated problems are also where our multigrid preconditioner has a larger speedup over the other methods.
We believe that this is because the multigrid preconditioner does a good job of capturing long range effects in the problem.

\subsection{Solver Accuracy}\label{sec:solver-accuracy}

\begin{figure}
  \centering
  \resizebox{!}{5cm}{\input{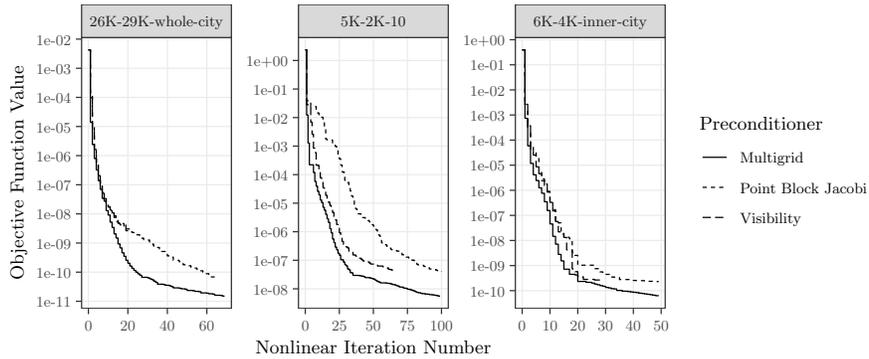}}
  \caption{
    Objective function value vs nonlinear iteration number for a variety of synthetic problems with varying problem structure.
    Our multigrid preconditioner tends to reach that value in fewer iterations than the other preconditioners because it is more accurate for a given solve tolerance.
    For solves where a high accuracy is required, or where Jacobian calculation is expensive, our preconditioner is a good choice.
  }
  \label{fig:cost-iteration}
\end{figure}


Our multigrid preconditioner is a more accurate preconditioner than point block Jacobi and visibility based methods.
In general, preconditioners like point block Jacobi converge fast in the residual norm, but converge slower in the error norm.
Multigrid tends to converge similarly in the error norm and the residual norm.
This behavior is reflected in the nonlinear convergence when using our preconditioner vs point block Jacobi.
Each nonlinear iteration with multigrid reduces the objective function by a larger value than point block Jacobi, indicating that the multigrid solution was more accurate.
See figure \ref{fig:cost-iteration} for plots of this behavior on some of our datasets.
Also interesting to note in this figure is the slope of convergence.
In almost all the plots, the solvers first converge quickly then hit a point where they start converging more slowly.
Our multigrid preconditioner also follows this characteristic, but converges more steeply in the first phase and continues converging quickly for longer.
We believe this is because our preconditioner more accurately captures long range effects.
For nonlinear optimization problems where a high degree of accuracy is required, this behavior makes our multigrid preconditioner even more performant than existing preconditioners.

For solves where $\tau$ is smaller (0.01), our preconditioner performs better than point block Jacobi.
When $\tau$ is larger, our preconditioner is generally slower than point block Jacobi because the setup cost of our preconditioner is not amortized.
In general, our preconditioner is a good choice when when tight (small) solve tolerances are used or when the linear problems are hard to solve.

\subsection{Scaling}

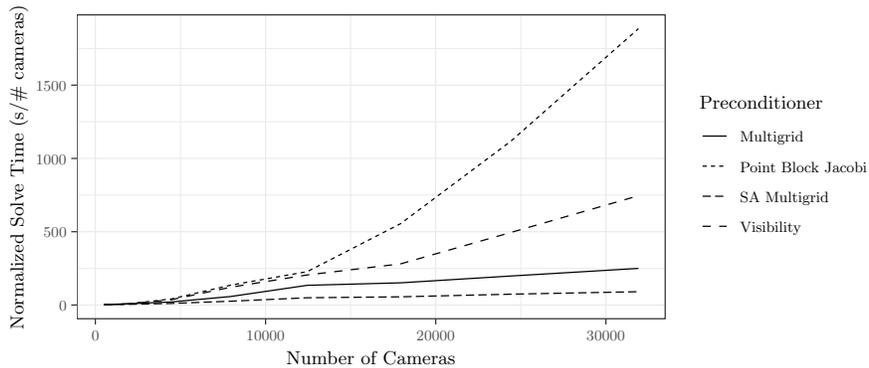
\begin{figure}
  \centering
  \resizebox{!}{5cm}{
\begin{tikzpicture}[x=1pt,y=1pt]
\definecolor{fillColor}{RGB}{255,255,255}
\path[use as bounding box,fill=fillColor,fill opacity=0.00] (0,0) rectangle (433.62,180.67);
\begin{scope}
\path[clip] (  0.00,  0.00) rectangle (433.62,180.67);
\definecolor{drawColor}{RGB}{255,255,255}
\definecolor{fillColor}{RGB}{255,255,255}

\path[draw=drawColor,line width= 0.6pt,line join=round,line cap=round,fill=fillColor] (  0.00,  0.00) rectangle (433.62,180.68);
\end{scope}
\begin{scope}
\path[clip] ( 38.03, 29.10) rectangle (320.57,175.17);
\definecolor{fillColor}{RGB}{255,255,255}

\path[fill=fillColor] ( 38.03, 29.10) rectangle (320.57,175.17);
\definecolor{drawColor}{gray}{0.92}

\path[draw=drawColor,line width= 0.3pt,line join=round] ( 38.03, 53.31) --
	(320.57, 53.31);

\path[draw=drawColor,line width= 0.3pt,line join=round] ( 38.03, 88.54) --
	(320.57, 88.54);

\path[draw=drawColor,line width= 0.3pt,line join=round] ( 38.03,123.78) --
	(320.57,123.78);

\path[draw=drawColor,line width= 0.3pt,line join=round] ( 38.03,159.01) --
	(320.57,159.01);

\path[draw=drawColor,line width= 0.3pt,line join=round] ( 87.74, 29.10) --
	( 87.74,175.17);

\path[draw=drawColor,line width= 0.3pt,line join=round] (169.48, 29.10) --
	(169.48,175.17);

\path[draw=drawColor,line width= 0.3pt,line join=round] (251.22, 29.10) --
	(251.22,175.17);

\path[draw=drawColor,line width= 0.6pt,line join=round] ( 38.03, 35.69) --
	(320.57, 35.69);

\path[draw=drawColor,line width= 0.6pt,line join=round] ( 38.03, 70.93) --
	(320.57, 70.93);

\path[draw=drawColor,line width= 0.6pt,line join=round] ( 38.03,106.16) --
	(320.57,106.16);

\path[draw=drawColor,line width= 0.6pt,line join=round] ( 38.03,141.39) --
	(320.57,141.39);

\path[draw=drawColor,line width= 0.6pt,line join=round] ( 46.86, 29.10) --
	( 46.86,175.17);

\path[draw=drawColor,line width= 0.6pt,line join=round] (128.61, 29.10) --
	(128.61,175.17);

\path[draw=drawColor,line width= 0.6pt,line join=round] (210.35, 29.10) --
	(210.35,175.17);

\path[draw=drawColor,line width= 0.6pt,line join=round] (292.09, 29.10) --
	(292.09,175.17);
\definecolor{drawColor}{RGB}{0,0,0}

\path[draw=drawColor,line width= 0.6pt,line join=round] ( 50.87, 35.86) --
	( 63.04, 36.33) --
	( 83.39, 37.16) --
	(111.91, 39.79) --
	(148.60, 45.16) --
	(193.47, 46.36) --
	(246.50, 49.64) --
	(307.73, 53.29);

\path[draw=drawColor,line width= 0.6pt,dash pattern=on 2pt off 2pt ,line join=round] ( 50.87, 35.75) --
	( 63.04, 36.28) --
	( 83.39, 38.62) --
	(111.91, 45.26) --
	(148.60, 51.72) --
	(193.47, 74.80) --
	(246.50,114.49) --
	(307.73,168.54);

\path[draw=drawColor,line width= 0.6pt,dash pattern=on 4pt off 2pt ,line join=round] ( 50.87, 35.85) --
	( 63.04, 36.03) --
	( 83.39, 36.46) --
	(111.91, 37.52) --
	(148.60, 39.18) --
	(193.47, 39.63) --
	(246.50, 40.90) --
	(307.73, 42.09);

\path[draw=drawColor,line width= 0.6pt,dash pattern=on 4pt off 4pt ,line join=round] ( 50.87, 35.74) --
	( 63.04, 36.26) --
	( 83.39, 38.25) --
	(111.91, 44.26) --
	(148.60, 50.15) --
	(193.47, 55.47) --
	(246.50, 70.59) --
	(307.73, 88.22);
\definecolor{drawColor}{gray}{0.20}

\path[draw=drawColor,line width= 0.6pt,line join=round,line cap=round] ( 38.03, 29.10) rectangle (320.57,175.17);
\end{scope}
\begin{scope}
\path[clip] (  0.00,  0.00) rectangle (433.62,180.67);
\definecolor{drawColor}{gray}{0.30}

\node[text=drawColor,anchor=base east,inner sep=0pt, outer sep=0pt, scale=  0.80] at ( 33.08, 32.94) {0};

\node[text=drawColor,anchor=base east,inner sep=0pt, outer sep=0pt, scale=  0.80] at ( 33.08, 68.17) {500};

\node[text=drawColor,anchor=base east,inner sep=0pt, outer sep=0pt, scale=  0.80] at ( 33.08,103.40) {1000};

\node[text=drawColor,anchor=base east,inner sep=0pt, outer sep=0pt, scale=  0.80] at ( 33.08,138.64) {1500};
\end{scope}
\begin{scope}
\path[clip] (  0.00,  0.00) rectangle (433.62,180.67);
\definecolor{drawColor}{gray}{0.20}

\path[draw=drawColor,line width= 0.6pt,line join=round] ( 35.28, 35.69) --
	( 38.03, 35.69);

\path[draw=drawColor,line width= 0.6pt,line join=round] ( 35.28, 70.93) --
	( 38.03, 70.93);

\path[draw=drawColor,line width= 0.6pt,line join=round] ( 35.28,106.16) --
	( 38.03,106.16);

\path[draw=drawColor,line width= 0.6pt,line join=round] ( 35.28,141.39) --
	( 38.03,141.39);
\end{scope}
\begin{scope}
\path[clip] (  0.00,  0.00) rectangle (433.62,180.67);
\definecolor{drawColor}{gray}{0.20}

\path[draw=drawColor,line width= 0.6pt,line join=round] ( 46.86, 26.35) --
	( 46.86, 29.10);

\path[draw=drawColor,line width= 0.6pt,line join=round] (128.61, 26.35) --
	(128.61, 29.10);

\path[draw=drawColor,line width= 0.6pt,line join=round] (210.35, 26.35) --
	(210.35, 29.10);

\path[draw=drawColor,line width= 0.6pt,line join=round] (292.09, 26.35) --
	(292.09, 29.10);
\end{scope}
\begin{scope}
\path[clip] (  0.00,  0.00) rectangle (433.62,180.67);
\definecolor{drawColor}{gray}{0.30}

\node[text=drawColor,anchor=base,inner sep=0pt, outer sep=0pt, scale=  0.80] at ( 46.86, 18.64) {0};

\node[text=drawColor,anchor=base,inner sep=0pt, outer sep=0pt, scale=  0.80] at (128.61, 18.64) {10000};

\node[text=drawColor,anchor=base,inner sep=0pt, outer sep=0pt, scale=  0.80] at (210.35, 18.64) {20000};

\node[text=drawColor,anchor=base,inner sep=0pt, outer sep=0pt, scale=  0.80] at (292.09, 18.64) {30000};
\end{scope}
\begin{scope}
\path[clip] (  0.00,  0.00) rectangle (433.62,180.67);
\definecolor{drawColor}{RGB}{0,0,0}

\node[text=drawColor,anchor=base,inner sep=0pt, outer sep=0pt, scale=  1.00] at (179.30,  7.44) {Number of Cameras};
\end{scope}
\begin{scope}
\path[clip] (  0.00,  0.00) rectangle (433.62,180.67);
\definecolor{drawColor}{RGB}{0,0,0}

\node[text=drawColor,rotate= 90.00,anchor=base,inner sep=0pt, outer sep=0pt, scale=  1.00] at ( 12.39,102.14) {Normalized Solve Time (s/\# cameras)};
\end{scope}
\begin{scope}
\path[clip] (  0.00,  0.00) rectangle (433.62,180.67);
\definecolor{fillColor}{RGB}{255,255,255}

\path[fill=fillColor] (331.57, 60.81) rectangle (428.12,143.46);
\end{scope}
\begin{scope}
\path[clip] (  0.00,  0.00) rectangle (433.62,180.67);
\definecolor{drawColor}{RGB}{0,0,0}

\node[text=drawColor,anchor=base west,inner sep=0pt, outer sep=0pt, scale=  1.00] at (337.07,130.10) {Preconditioner};
\end{scope}
\begin{scope}
\path[clip] (  0.00,  0.00) rectangle (433.62,180.67);
\definecolor{fillColor}{RGB}{255,255,255}

\path[fill=fillColor] (337.07,109.67) rectangle (351.52,124.13);
\end{scope}
\begin{scope}
\path[clip] (  0.00,  0.00) rectangle (433.62,180.67);
\definecolor{drawColor}{RGB}{0,0,0}

\path[draw=drawColor,line width= 0.6pt,line join=round] (338.52,116.90) -- (350.08,116.90);
\end{scope}
\begin{scope}
\path[clip] (  0.00,  0.00) rectangle (433.62,180.67);
\definecolor{fillColor}{RGB}{255,255,255}

\path[fill=fillColor] (337.07, 95.22) rectangle (351.52,109.67);
\end{scope}
\begin{scope}
\path[clip] (  0.00,  0.00) rectangle (433.62,180.67);
\definecolor{drawColor}{RGB}{0,0,0}

\path[draw=drawColor,line width= 0.6pt,dash pattern=on 2pt off 2pt ,line join=round] (338.52,102.45) -- (350.08,102.45);
\end{scope}
\begin{scope}
\path[clip] (  0.00,  0.00) rectangle (433.62,180.67);
\definecolor{fillColor}{RGB}{255,255,255}

\path[fill=fillColor] (337.07, 80.77) rectangle (351.52, 95.22);
\end{scope}
\begin{scope}
\path[clip] (  0.00,  0.00) rectangle (433.62,180.67);
\definecolor{drawColor}{RGB}{0,0,0}

\path[draw=drawColor,line width= 0.6pt,dash pattern=on 4pt off 2pt ,line join=round] (338.52, 87.99) -- (350.08, 87.99);
\end{scope}
\begin{scope}
\path[clip] (  0.00,  0.00) rectangle (433.62,180.67);
\definecolor{fillColor}{RGB}{255,255,255}

\path[fill=fillColor] (337.07, 66.31) rectangle (351.52, 80.77);
\end{scope}
\begin{scope}
\path[clip] (  0.00,  0.00) rectangle (433.62,180.67);
\definecolor{drawColor}{RGB}{0,0,0}

\path[draw=drawColor,line width= 0.6pt,dash pattern=on 4pt off 4pt ,line join=round] (338.52, 73.54) -- (350.08, 73.54);
\end{scope}
\begin{scope}
\path[clip] (  0.00,  0.00) rectangle (433.62,180.67);
\definecolor{drawColor}{RGB}{0,0,0}

\node[text=drawColor,anchor=base west,inner sep=0pt, outer sep=0pt, scale=  0.80] at (356.52,114.15) {Multigrid};
\end{scope}
\begin{scope}
\path[clip] (  0.00,  0.00) rectangle (433.62,180.67);
\definecolor{drawColor}{RGB}{0,0,0}

\node[text=drawColor,anchor=base west,inner sep=0pt, outer sep=0pt, scale=  0.80] at (356.52, 99.69) {Point Block Jacobi};
\end{scope}
\begin{scope}
\path[clip] (  0.00,  0.00) rectangle (433.62,180.67);
\definecolor{drawColor}{RGB}{0,0,0}

\node[text=drawColor,anchor=base west,inner sep=0pt, outer sep=0pt, scale=  0.80] at (356.52, 85.24) {SA Multigrid};
\end{scope}
\begin{scope}
\path[clip] (  0.00,  0.00) rectangle (433.62,180.67);
\definecolor{drawColor}{RGB}{0,0,0}

\node[text=drawColor,anchor=base west,inner sep=0pt, outer sep=0pt, scale=  0.80] at (356.52, 70.78) {Visibility};
\end{scope}
\end{tikzpicture}}
  \caption{
    Preconditioned linear solver scaling experiment on a series of increasingly larger grids with long range noise only.
    Grid size is on the order of $\sqrt{\text{number of cameras}} \times \sqrt{\text{number of cameras}}$.
    The y-axis is a measure of linear solver solve time (not including linear solver setup) per camera.
    A horizontal trend indicate that a solver is scaling linearly with the number of cameras.
    Slopes greater than zero indicates the solver is scaling superlinearly.
    We see the expected behavior that Multigrid scales close to linearly while visibility and point block Jacobi scale superlinearly.
    Smoothed aggregation multigrid has the best scaling, but its setup phase is prohibitively expensive.
  }
  \label{fig:scaling-test}
\end{figure}

For larger problem sizes, the algorithmic complexity of different solution techniques begins to dominate over constant factors.
It is well known that solving a second order elliptic system (such as elasticity) on a $\sqrt n\times \sqrt n$ grid using conjugate gradients with point block Jacobi preconditioning requires $O(n^{1/2})$ iterations, for a total cost of $O(n^{1.5})$ \cite{toselli2005domain}.
We expect to interpret the global coupling and scaling of bundle adjustment similarly, in terms of diameter of the visibility graph, which has 2D grid structure for street view data in cities.
If the structure is not two dimensional, say for a long country road, then we would expect the bound to be $O(n^2)$.
We expect that visibility based methods also scale as $O(n^{1.5})$, but with different constant factors as they cannot handle long range effects.
Multigrid can be bounded by $O(n)$, but this requires certain conditions on the prolongation operator that we do not satisfy (specifically, not smoothing the prolongation operator means that we do not satisfy the strong approximation property).
Empirically, we find that our multigrid technique does not scale linearly with problem size, but still scales better than other preconditioners.

To empirically verify the scaling of visibility-based methods and our multigrid method, we construct a series of city block-like problems with increasing numbers of blocks.
Increasing the number of city blocks instead of adding more cameras to the same structure means that the test problems have increasing diameter.
We add noise that looks like a sin wave to the problem to induce long range errors.
Figure~\ref{fig:scaling-test} shows the results of this experiment.
Surprisingly, point block Jacobi is scaling as $O(n^2)$, which indicates that bundle adjustment is more similar to a shell problem than an elasticity problem.

\begin{figure}
  \centering
  \resizebox{!}{5cm}{\input{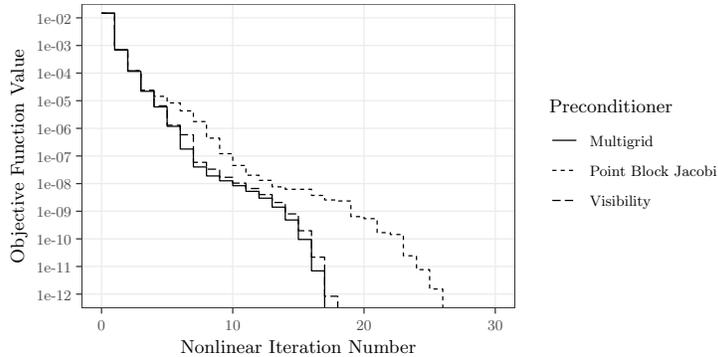}}
  \caption{
    Objective function value vs nonlinear iteration number on a 40 by 40 synthetic city grid.
    The linear solve tolerance is $\tau=0.01$ with a Huber loss function to expose long range error.
    The objective function value for solves with point block Jacobi lags behind solves with multigrid or visibility preconditioners because point block Jacobi is a less accurate preconditioner.
  }
  \label{fig:iteration-counts}
\end{figure}

\subsection{Parallelism}

Our solver is currently single threaded.
Most of the time spent in the solver is in linear algebra, so using a parallel linear algebra framework is an easy way to parallelism.
The only non-linear algebra part of our multigrid solver is aggregation.
There are parallel aggregation techniques, but they require more work than simply swapping out a library.

\subsection{Robust Error Metrics}\label{sec:robust}

Often, there are outlying points in bundle adjustment problems.
These are the product of incorrect correspondences, points that are too close to accurately track, or points with very poor initialization.
In any case, outlying points make up a disproportionate amount of the objective function (due to the quadratic scaling of the reprojection error).
Levenberg-Marquardt attempts to minimize the error, and the quickest way it does is to fix each outlier in turn.
This effectively masks the presence of long-range error, leading to good linear solver performance but poor nonlinear convergence.
The usual solution is to use a robust loss function.
Robust loss functions are quadratic around the origin, but become linear the farther they get from the origin.
The robust loss function we use is Huber loss,
$$
\text{loss}(x) = \begin{cases}
                   x & x \leq 1, \\
                   2\sqrt{x}-1 & x > 1,
                 \end{cases}
$$
where $x$ is the squared L2 norm of the residuals.

Point block Jacobi is a local preconditioner: it is effective at resolving noise in a small neighborhood.
Without a robust loss function point block Jacobi is quick because it ``fixes'' outliers in a couple iterations.
A robust loss function exposes long-range noise making point block Jacobi slow.
However, multigrid is more effective at addressing long range error, so it is a comparatively faster preconditioner when used with a robust loss function.

%

\section{Conclusion \& Future Work}

We present a multigrid preconditioner for conjugate gradients that performs better than any existing preconditioner and solver on bundle adjustment problems with long range effects or problems requiring a high solve tolerance.
In tests on a set of large synthetic problems, our preconditioner is up to 13 times faster than the next best preconditioner.
Our preconditioner is tailored for a specific kind of bundle adjustment problem: a 9-parameter camera model with reprojection error.
Generalizing this preconditioner to different kinds of camera models would require computing a new analytical nullspace.
For most models, this should just involve finding the instantaneous derivatives of the 7 free modes (3x translation, 3x rotation, 1x scaling).
It would also be possible to use an eigensolver to find the near-nullspace at an increase in setup time cost.

In future work we would like to find a way to automatically switch between point block Jacobi and multigrid preconditioners depending on the difficultly of the linear problem.
We would also like to improve our preconditioner so that it will scale linearly with the problem size by either using some sort of filtered smoothing, or other multigrid techniques used to compensate for lack of prolongation smoothing.

\clearpage

\section*{Acknowledgments}
This material is based upon work supported by the U.S. Department of Energy, Office of Science, Office of Advanced Scientific Computing Research under Award Number DE-SC0016140.
This research used resources of the National Energy Research Scientific Computing Center, a DOE Office of Science User Facility supported by the Office of Science of the U.S. Department of Energy under Contract No. DE-AC02-05CH11231.

\bibliographystyle{splncs04}
\bibliography{refs}

\end{document}